%% file: iclr2020_conference.tex
\documentclass{article} 
\usepackage{iclr2020_conference,times}
\newcommand{\todo}[1]{}
\renewcommand{\todo}[1]{{\color{red} TODO: {#1}}}

\input{math_commands.tex}

\usepackage{hyperref}
\usepackage{url}
\usepackage{graphicx}
\usepackage{cleveref}
\usepackage{multicol}
\usepackage{multirow}
\usepackage{amsmath}
\usepackage{tabularx}
\usepackage{pbox}
\usepackage{latexsym}
\usepackage{booktabs}
\usepackage{varwidth}
\usepackage{caption}

\title{Progressive Growing of Neural ODEs}

\author{Hammad A. Ayyubi \thanks{This work was done while the author was interning at SRI International.} \\
University of California, San Diego\\
\texttt{hayyubi@ucsd.edu} \\
\And
Yi Yao \& Ajay Divakaran \\
SRI International \\
\texttt{\{yi.yao,ajay.divakaran\}@sri.com} \\
}

\iclrfinalcopy 
\begin{document}
\vspace{-10cm}
\maketitle
\vspace{-12pt}
\begin{abstract}
\vspace{-7pt}
Neural Ordinary Differential Equations (NODEs) have proven to be a powerful modeling tool for approximating (interpolation) and forecasting (extrapolation) irregularly sampled time series data. However, their performance degrades substantially when applied to real-world data, especially long-term data with complex behaviors (e.g., long-term trend across years, mid-term seasonality across months, and short-term local variation across days). To address the modeling of such complex data with different behaviors at different frequencies (time spans), we propose a novel progressive learning paradigm of NODEs for long-term time series forecasting. Specifically, following the principle of curriculum learning, we gradually increase the complexity of data and network capacity as training progresses. Our experiments with both synthetic data and real traffic data (PeMS Bay Area traffic data) show that our training methodology consistently improves the performance of vanilla NODEs by over 64\%.
\end{abstract}

\vspace{-10pt}
\section{Introduction}
\vspace{-10pt}
Time series analysis is critical in a number of domains such as stock prices analysis, weather analysis, business planning, resource allocation, etc. One major aspect of such time series analysis is dealing with irregularly sampled data. Prior approaches tackle this issue by mapping such data onto equally spaced intervals \citep{pmlr-v56-Lipton16}. However, this approximation introduces error, especially at the local maxima and minima of seasonal fluctuations.

Several approaches have been proposed to improve approximation accuracy. \citet{mei2016neural} uses exponential decay to model state between observations. Neural Ordinary Differential Equations (NODEs) \cite{chen2018neural} model continuous states between observations using a continuous depth black box ODE solver parameterized by a neural network. NODEs have proven to be promising for forecasting problems with irregular samples \citep{rubanova2019latent}. However, they are brittle when tasked with forecasting functions containing long-term trends (yearly) and short-term seasonalities (monthly or daily).

To address the above mentioned issue, we propose novel networks based on Progressive Neural Ordinary Equations (PODEs). Specifically, we follow a curriculum learning approach in which we gradually increase the data complexity as well as network complexity as training progresses. The key idea is that the network learns low frequency and easier to learn trends first and then the high frequency and more complex seasonalities. Such a breakdown of task enables the network to gradually learn these complex curves, which is, otherwise, too difficult to learn.

We summarize the contribution of the paper as follows:

\begin{itemize}
    \item We propose novel Progressive Neural ODEs (PODEs) for the analysis of irregularly sampled complex time series containing trends and seasonalities.
    \item We demonstrate empirical evidence of the superiority of our approach as compared to vanilla NODEs on both synthetic and real-world data.
\end{itemize}

\section{Related Work}
\vspace{-10pt}
\paragraph{Time Series Modeling}
Compared with the extensive body of work on time series forecasting of regularly sampled (i.e., equally-spaced) data (\cite{box2015, brockwell2016}), fewer methods exist  for irregularly sampled (i.e., unevenly-spaced) data. Analysis of such data becomes a critical challenge associated with complex real-world applications such as economics (\cite{econ_uneven}), healthcare (\cite{health_uneven}), and astronomy (\cite{astro_uneven}), to name a few. One major line of methods transform irregularly spaced samples into equally spaced ones and then apply existing methods for equally spaced data (\cite{arima}). For instance,  Gaussian Process combined with learned neural networks is recently applied for interpolating irregularly sampled data (\cite{health_uneven, int_networks}). However, such methods suffer from a number of biases (\cite{rehfeld2011}), which significantly degrades the overall performance, especially for highly irregular observations. Classical exponential smoothing methods such as Holt and Winters (\cite{hw, holt2004}) are applied to irregularly-sampled time series mainly for the estimation of trends and seasonals. With the significant development in deep learning, learned networks in a data-driven manner (e.g., NODEs) also find promising applications to analysis of irregularly sampled data.

\paragraph{Curriculum Learning}
\vspace{-10pt}
The idea that humans learn in an organized manner, leveraging previous experience/knowledge to learn more complex tasks, originates from cognitive science. \citet{Elman1993LearningAD} first explored whether this idea can be used to train neural networks. He showed how curriculum learning can be used to learn simple language grammar. \citet{bengio16curlearn} extended this idea and showed it's effectiveness in language modeling and geometric shape recognition task. More recently, this idea of progressive learning has been explored in many contexts. \citet{ZarembaS14} use this approach for evaluating short computer programs. \citet{MatiisenOCS17} apply curriculum learning in reinforcement learning regime, where a student learns a complex task following a teacher's direction of learning subtasks. \citet{karras2017progressive} show impressive performance in generating human faces using Generative Adversarial Networks (GANs) following the same learning strategy. Inspired by this line of works, we plan to apply progressive learning strategy to NODE-based time series forecasting in order to improve its performance on complex real-world data with trends and seasonalities.
\vspace{-5pt}
\section{Our Approach}

\begin{figure}[t]
\includegraphics[width=\linewidth]{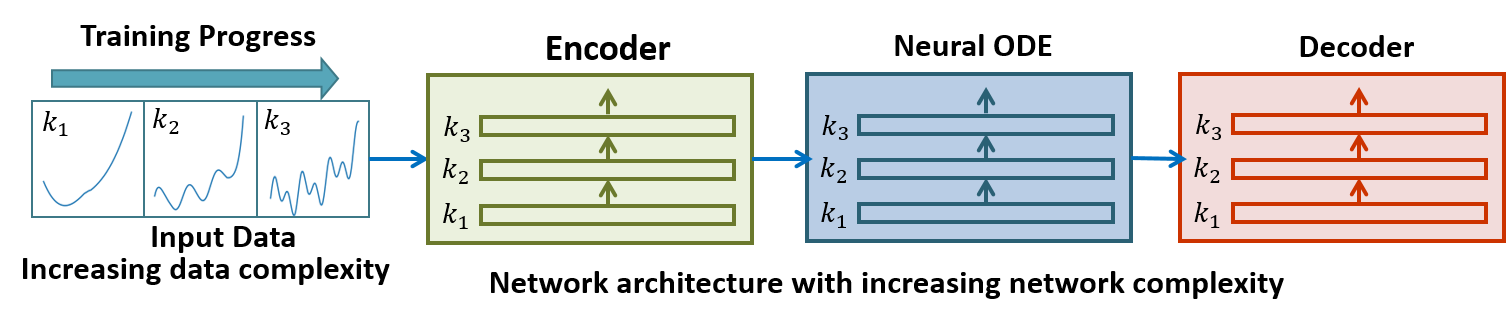}
\caption{Overview of our approach. Following curriculum learning approach, data complexity  and network complexity is gradually increased (from $k_1$ to $k_3$) as training progresses.}
\label{fig:model}
\end{figure}

\vspace{-10pt}
To illustrate the advantages of PODEs, we choose to use the same network design originally proposed in \citet{rubanova2019latent} as our backbone architecture. This architecture consists of an encoder, a NODE, and a decoder as shown in \cref{fig:model}. The encoder maps the input data into a fixed length embedding. The NODE network models the temporal dynamics of the data using irregularly spaced samples and make prediction of future values. Finally, the decoder transforms the prediction represented in the latent embedding space to actual output.

The fact that this architecture, despite it's theoretical advantage, failed to model complex time-series functions containing trends and seasonalities (section \ref{sec:syn} and \ref{sec:real}), prompted us to employ a progressive learning approach. Under this scheme, we reorganize network layers into groups and train each group of layers progressively using data with gradually increasing complexity. The key idea is that we divide the complex task of learning functions containing trends and seasonalities into much easier to learn sub-tasks.

The network architecture and training procedure is illustrated in \cref{fig:model}. We divide the training stages into $k$ steps. At each step, we add a group of layers to the encoder, NODE and decoder. Concurrently, at each step we increase the complexity of input data. All the network layers - both the previously trained and the newly added ones - remain trainable throughout training. To alleviate instability introduced by adding new layers, we use alpha blending - gradual addition of the new layer controlled by the parameter $\alpha$. We prepare the input data for the $1, ..., k-1$ steps using $k-1$ low pass filters. Note that the original data is used as the input for the $k^{th}$ step. 

\begin{figure*}[t]
    \includegraphics[width=\textwidth]{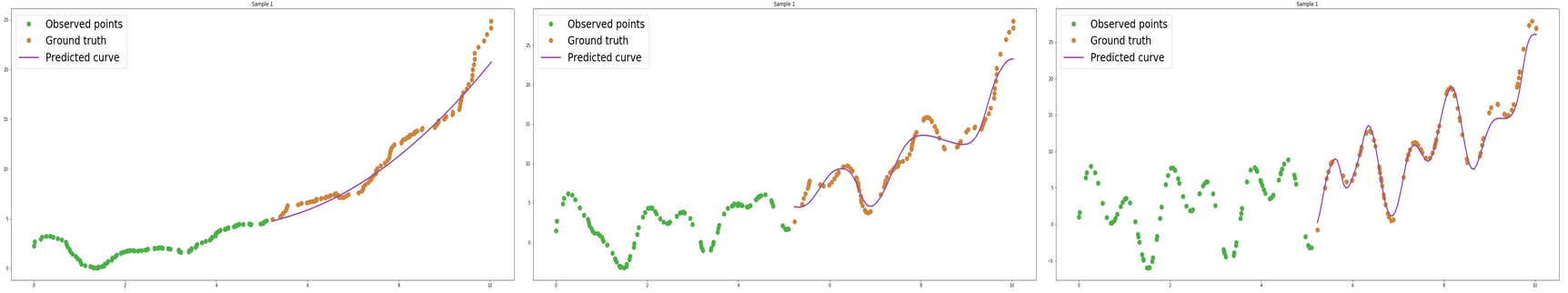}\hfill
    \caption{Forecasting performance at different training stages. Left: $k=1$. Middle: $k=2$. Right: $k=3$. As training progresses, data complexity along with network complexity increases enabling the learning of complex time series.Green dots: irregularly sampled observations. Orange dots: ground truth. Purple curve: prediction.}
    \label{fig:progress}
\end{figure*}

\begin{figure}[t]
\includegraphics[width=\linewidth]{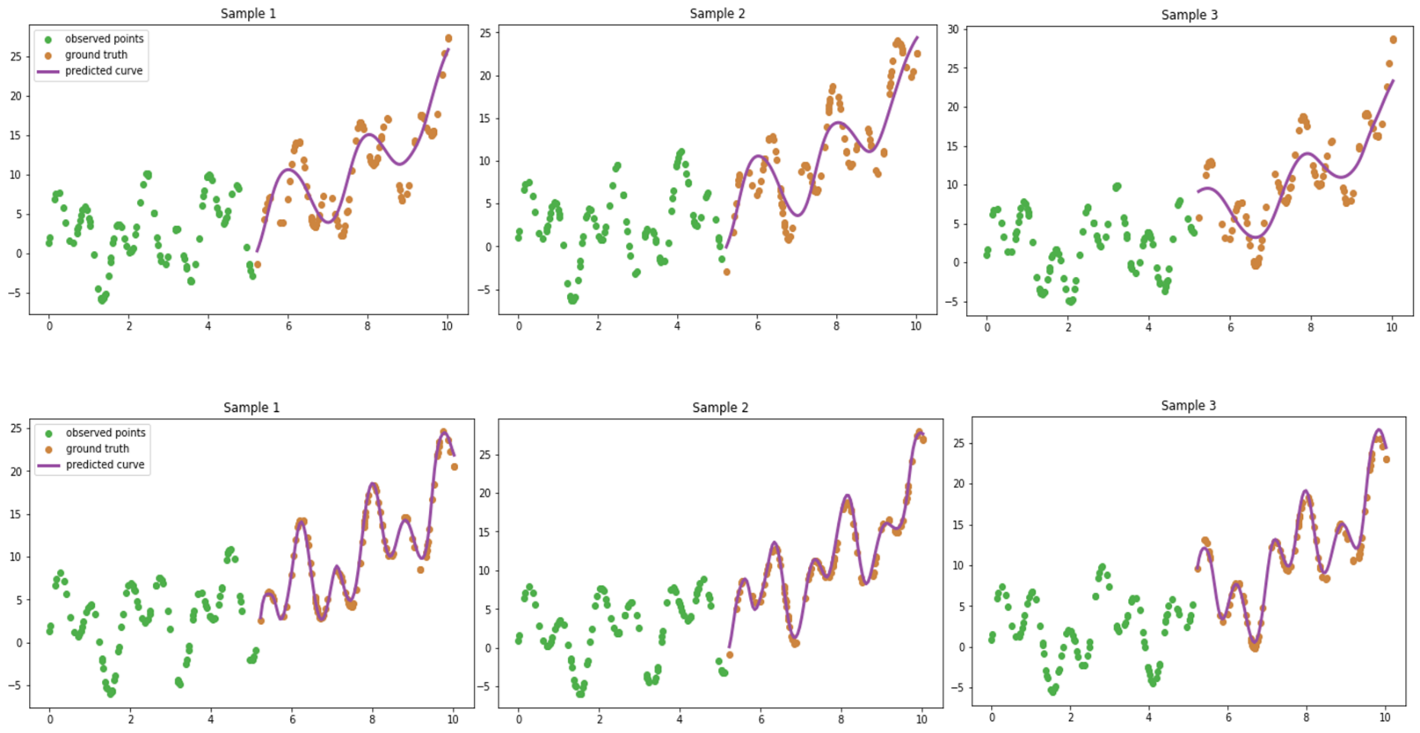}
\caption{Forecasting performance comparison between NODE (top) and PODE (bottom) on synthetic data - $exp(cx) + sin(t_1x) + sin(t_2x)$. Each column shows samples with different $c$ for trends and $t_1$, $t_2$ for seasonal fluctuations. Overall speaking, NODE can capture trend ($c$) and one seasonal  component with lower frequency ($t_1$) whereas PODE can capture not only trend ($c$) but also both seasonalities ($t_1$ and $t_2$). Green dots: irregularly sampled observations. Orange dots: ground truth. Purple curve: prediction.}
\label{fig:syn_results}
\end{figure}

\section{Experiments}
\vspace{-10pt}
We use a Gated Recurrent Unit as encoder and a feed forward network as the decoder. We train our network in $k=3$ steps. At each step, we add a layer. Therefor, we have three layers in total in each of our sub networks. To be consistent with NODEs, we use the same hyperparameters as described in \citet{rubanova2019latent} - an initial learning rate of 1e-2 with an exponentially decaying schedule and a batch size of 50. 

We compare the performance of PODE with respect to the original NODE \citet{rubanova2019latent} using two datasets: a synthetic datasets for experiments with controlled parameters such as trend rate and seasonal frequencies and a real-world dataset - the PeMS-Bay traffic datasets, a commonly used dataset for time series forecasting. We also compare our model to traditional approaches: (1) Static Model: predicts the same value as encountered $p$ time steps before, (2) Historical Average (HA): predicts weighted average of past seasons as its forecast, and (3) ARIMA: Auto-Regressive Integrated Moving Average which is an auto-regressive model popularly used for time series prediction.

\subsection{Synthetic Data}
\label{sec:syn}
\vspace{-10pt}

We generate synthetic data of the form $exp(cx)+ sin(t_1x)+ sin(t_2x)$, where $c$, $t_1$, $t_2$ control the global trend and local seasonals. This is a simplified version of real-world data with trends (i.e., the $exp$ function) and multiple seasonalities (i.e., the two $sin$ functions). We vary the $exp$ and $sin$ constants ($c$, 
$t_1$, $t_2$) and add Gaussian noise to the sampled points. In total, we generate $1000$ example and use $80\%$-$20\%$ train-test split. For each example, we irregularly sample 200 points- showing the network a fraction of these points (i.e., the first 100 samples) and asking it to forecast the rest (i.e., the remaining 100 samples).

\Cref{fig:progress} illustrates the prediction performance at different training stages with increased data and network complexity. We can see that by breaking down the complex curve into simpler curves, our PODE can learn complex functions incrementally.  \Cref{fig:syn_results} compares the forecasting performance between NODEs and PODEs. NODEs fail to represent the complex dynamics of our synthetic curves, yielding a mean squared error (MSE) of 15.56. In comparison, PODEs are capable of capturing both trend and seasonalities, producing an MSE Of 0.81, a substantial improvement. \Cref{tab:syn_results} compares the MSE of PODE against other baselines.


\begin{table}[t]
\small
\centering
  \begin{tabular}{lccccc}
    \toprule
    \multirow{2}{*}[-0.5\dimexpr \aboverulesep + \belowrulesep + \cmidrulewidth]{\textbf{Dataset}}
    & \multicolumn{5}{c}{\textbf{Models}}\\
    \cmidrule(l){2-6}
    & Static \hspace{-2mm} & HA \hspace{-2mm} & ARIMA \hspace{-2mm} & NODE \hspace{-2mm} & PODE \hspace{-2mm}\\
    \midrule
    Synthetic & 36.43 & 35.74 & 29.69 & 15.56 & \textbf{0.81} \\
    PEMS-BAY (+E03) &  17.58  & 40.45 & 47.78 & 13.80 & \textbf{4.87} \\
    \bottomrule
  \end{tabular}
  \caption{MSE on synthetic and PEMS-BAY datasets.
  }
  \label{tab:syn_results}
\end{table}

\subsection{PeMS Bay Area Traffic Data}
\label{sec:real}
\vspace{-10pt}
This dataset is collected by California Department of Transportation (Caltrans) using Caltrans Performance Measurement System (PeMS). We use the traffic flow readings - average vehicles on per unit time - aggregated in five minutes interval. This dataset incorporates both trend and seasonalities (including weekly and daily changes) and, therefore, is commonly used for time series forecasting. It also contains sufficient equally-spaced samples, which allow us to conduct experiments on the effect of irregular spacing with different sampling strategy (e.g., maximum/minimum spaces).

\begin{figure}[t]
\includegraphics[width=\linewidth]{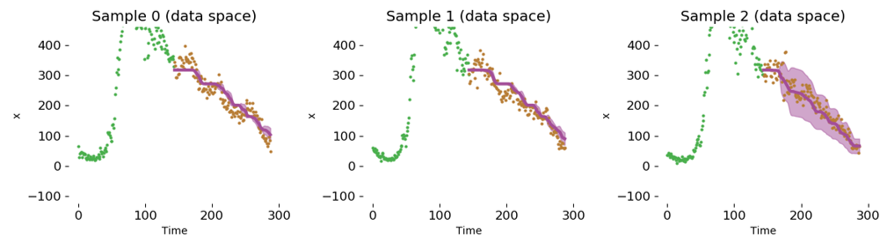}
\setlength{\belowcaptionskip}{-10pt}
\caption{Forecasting performance PODE on PeMS-BAY. Each column shows samples from different sensors. Green dots: irregularly sampled observations. Orange dots: ground truth. Purple curve: prediction. Purple shaded area: prediction uncertainty.}
\label{fig:real_results}
\end{figure}

We randomly select one of the sensor data to be used, over a three year period from January, 2014 to December, 2017. Each sample in our data is a daily measurement of flow readings. This gives us 932 samples, after cleaning up days on which the sensor didn't collect any data, and 288 data points in each sample. As with synthetic data, we divide the dataset in 80\%-20\% train-test split ratio. We input half of 288 points to the network and ask it to forecast the latter half. \Cref{tab:syn_results} shows that PODE improves over NODE by more than 64\%. Qualitative samples are visualized in \cref{fig:real_results}.

\vspace{-8pt}
\section{Conclusion}
\vspace{-10pt}
We proposed a novel progressive learning approach for modeling irregularly sampled time series data with complex trends and seasonalities. We demonstrated substantial improvements over state-of-the-art NODEs on both synthetic and real-world data. Our empirical study suggests that with the same architecture, network performance can be further improved by appropriate design of training procedure, such as curriculum learning, especially for complex tasks such as the forecasting of irregularly sampled time series with trends and seasonalities.


\subsubsection*{Acknowledgments}
This research is based upon work supported in part by the Office of the Director of National Intelligence (ODNI), Intelligence Advanced Research Projects Activity (IARPA), via 2018-18050400004. The views and conclusions contained herein are those of the authors and should not be interpreted as necessarily representing the official policies, either expressed or implied, of ODNI, IARPA, or the U.S. Government. The U.S. Government is authorized to reproduce and distribute reprints for governmental purposes notwithstanding any copyright annotation therein.
\vspace{-5pt}
\bibliography{iclr2020_conference}
\bibliographystyle{iclr2020_conference}

\end{document}

%% file: math_commands.tex
\usepackage{amsmath,amsfonts,bm}

\def\eqref#1{equation~\ref{#1}}

\def\1{\bm{1}}

\DeclareMathAlphabet{\mathsfit}{\encodingdefault}{\sfdefault}{m}{sl}
\SetMathAlphabet{\mathsfit}{bold}{\encodingdefault}{\sfdefault}{bx}{n}













%% file: iclr2020_conference.bbl
\begin{thebibliography}{19}
\providecommand{\natexlab}[1]{#1}
\providecommand{\url}[1]{\texttt{#1}}
\expandafter\ifx\csname urlstyle\endcsname\relax
  \providecommand{\doi}[1]{doi: #1}\else
  \providecommand{\doi}{doi: \begingroup \urlstyle{rm}\Url}\fi

\bibitem[Bengio et~al.(2009)Bengio, Louradour, Collobert, and
  Weston]{bengio16curlearn}
Yoshua Bengio, J\'{e}r\^{o}me Louradour, Ronan Collobert, and Jason Weston.
\newblock Curriculum learning.
\newblock In \emph{Proceedings of the 26th Annual International Conference on
  Machine Learning}, ICML ’09, pp.\  41–48, New York, NY, USA, 2009.
  Association for Computing Machinery.
\newblock ISBN 9781605585161.
\newblock \doi{10.1145/1553374.1553380}.
\newblock URL \url{https://doi.org/10.1145/1553374.1553380}.

\bibitem[Box et~al.(2015)Box, Jenkins, C., and Ljung]{box2015}
George E.~P. Box, Gwilym~M. Jenkins, Reinsel~Gregory C., and Greta~M Ljung.
\newblock \emph{Time Series Analysis: Forecasting and Control}, volume~1.
\newblock Wiley Series in Probability and Statistics, 2015.

\bibitem[Brockwell \& Davis(2016)Brockwell and Davis]{brockwell2016}
Peter~J. Brockwell and Richard~A. Davis.
\newblock \emph{Introduction to Time Series and Forecasting}, volume~1.
\newblock Springer, 2016.

\bibitem[Chen et~al.(2018)Chen, Rubanova, Bettencourt, and
  Duvenaud]{chen2018neural}
Ricky T.~Q. Chen, Yulia Rubanova, Jesse Bettencourt, and David Duvenaud.
\newblock Neural ordinary differential equations, 2018.

\bibitem[Elman(1993)]{Elman1993LearningAD}
Jeffrey~L. Elman.
\newblock Learning and development in neural networks: the importance of
  starting small.
\newblock \emph{Cognition}, 48:\penalty0 71--99, 1993.

\bibitem[Gardner~Jr(1985)]{hw}
Everette~S Gardner~Jr.
\newblock Exponential smoothing: The state of the art.
\newblock \emph{Journal of forecasting}, 4\penalty0 (1):\penalty0 1--28, 1985.

\bibitem[Harvey \& Todd(1983)Harvey and Todd]{econ_uneven}
Andrew~C Harvey and PHJ Todd.
\newblock Forecasting economic time series with structural and box-jenkins
  models: A case study.
\newblock \emph{Journal of Business \& Economic Statistics}, 1\penalty0
  (4):\penalty0 299--307, 1983.

\bibitem[Holt(2004)]{holt2004}
Charles~C. Holt.
\newblock Forecasting seasonals and trends by exponentially weighted moving
  averages.
\newblock \emph{International Journal of Forecasting}, 20:\penalty0 5--10,
  2004.

\bibitem[Karras et~al.(2017)Karras, Aila, Laine, and
  Lehtinen]{karras2017progressive}
Tero Karras, Timo Aila, Samuli Laine, and Jaakko Lehtinen.
\newblock Progressive growing of gans for improved quality, stability, and
  variation, 2017.

\bibitem[Li \& Marlin(2016)Li and Marlin]{health_uneven}
Steven Cheng-Xian Li and Benjamin~M Marlin.
\newblock A scalable end-to-end gaussian process adapter for irregularly
  sampled time series classification.
\newblock In \emph{Advances in neural information processing systems}, pp.\
  1804--1812, 2016.

\bibitem[Lipton et~al.(2016)Lipton, Kale, and Wetzel]{pmlr-v56-Lipton16}
Zachary~C Lipton, David Kale, and Randall Wetzel.
\newblock Directly modeling missing data in sequences with rnns: Improved
  classification of clinical time series.
\newblock In Finale Doshi-Velez, Jim Fackler, David Kale, Byron Wallace, and
  Jenna Wiens (eds.), \emph{Proceedings of the 1st Machine Learning for
  Healthcare Conference}, volume~56 of \emph{Proceedings of Machine Learning
  Research}, pp.\  253--270, Children's Hospital LA, Los Angeles, CA, USA,
  18--19 Aug 2016. PMLR.
\newblock URL \url{http://proceedings.mlr.press/v56/Lipton16.html}.

\bibitem[Matiisen et~al.(2017)Matiisen, Oliver, Cohen, and
  Schulman]{MatiisenOCS17}
Tambet Matiisen, Avital Oliver, Taco Cohen, and John Schulman.
\newblock Teacher-student curriculum learning.
\newblock \emph{CoRR}, abs/1707.00183, 2017.
\newblock URL \url{http://arxiv.org/abs/1707.00183}.

\bibitem[Mei \& Eisner(2016)Mei and Eisner]{mei2016neural}
Hongyuan Mei and Jason Eisner.
\newblock The neural hawkes process: A neurally self-modulating multivariate
  point process, 2016.

\bibitem[Rehfeld et~al.(2011)Rehfeld, Marwan, Heitzig, and Kurths]{rehfeld2011}
Kira Rehfeld, Norbert Marwan, Jobst Heitzig, and Juergen Kurths.
\newblock Comparison of correlation analysis techniques for irregularly sampled
  time series.
\newblock \emph{Nonlinear Processes in Geophysics}, 18:\penalty0 389--404,
  2011.

\bibitem[Rubanova et~al.(2019)Rubanova, Chen, and Duvenaud]{rubanova2019latent}
Yulia Rubanova, Ricky T.~Q. Chen, and David Duvenaud.
\newblock Latent odes for irregularly-sampled time series, 2019.

\bibitem[Scargle(1982)]{astro_uneven}
Jeffrey~D Scargle.
\newblock Studies in astronomical time series analysis. ii-statistical aspects
  of spectral analysis of unevenly spaced data.
\newblock \emph{The Astrophysical Journal}, 263:\penalty0 835--853, 1982.

\bibitem[Shukla \& Marlin(2018)Shukla and Marlin]{int_networks}
Satya~Narayan Shukla and Benjamin Marlin.
\newblock Interpolation-prediction networks for irregularly sampled time
  series.
\newblock 2018.

\bibitem[Zaremba \& Sutskever(2014)Zaremba and Sutskever]{ZarembaS14}
Wojciech Zaremba and Ilya Sutskever.
\newblock Learning to execute.
\newblock \emph{CoRR}, abs/1410.4615, 2014.
\newblock URL \url{http://arxiv.org/abs/1410.4615}.

\bibitem[Zhang(2003)]{arima}
G~Peter Zhang.
\newblock Time series forecasting using a hybrid arima and neural network
  model.
\newblock \emph{Neurocomputing}, 50:\penalty0 159--175, 2003.

\end{thebibliography}
